\documentclass[]{TEAI}

\usepackage{helvet}
\usepackage[utf8]{inputenc}
\usepackage[T1]{fontenc}
\usepackage{amsmath,amssymb,mathtools,amsfonts}
\usepackage{graphicx}
\usepackage{caption}
\usepackage{booktabs}
\usepackage{multirow}
\usepackage{xcolor}
\usepackage{colortbl}
\usepackage{hyperref}
\usepackage{url}
\usepackage{natbib}

\providecommand{\email}[1]{\href{mailto:#1}{#1}}
\colorlet{highlight}{cyan!10}
\definecolor{gaincolor}{RGB}{0,128,0}
\definecolor{losscolor}{RGB}{180,60,60}
\definecolor{bestcolor}{RGB}{255,245,220}
\definecolor{oraclecolor}{RGB}{95,95,95}
\newcommand{\gain}[1]{{\scriptsize\textcolor{gaincolor}{(+\,#1)}}}

\newcommand{\bestcell}[2]{\cellcolor{bestcolor}\textbf{#1}\,\gain{#2}}
\newcommand{\secondcell}[2]{\underline{#1}\,\gain{#2}}
\title{Disagree to Explore, Agree to Commit: Routing-Guided Test-Time Scaling for Software Agents}

\author{
Kang Chen\textsuperscript{1*},
Junjie Nian\textsuperscript{1*},
Yixin Cao\textsuperscript{1,2$\dagger$},
Yu-Gang Jiang\textsuperscript{1}
}

\affiliation[1]{\mbox{Fudan University}}
\affiliation[2]{\mbox{Shanghai Innovation Institute}}

\correspondence{\email{kchen24@m.fudan.edu.cn}, \email{yxcao@fudan.edu.cn}}
\checkdata[Web]{\href{https://CckFdu.com/RISA}{\texttt{https://CckFdu.com/RISA}}}

\abstract{
Software-engineering agents solve repository-level tasks through long, stochastic
tool-use trajectories, and repeated attempts often find fixes missed by one run.
Test-time scaling is difficult because patches lack canonical answer forms, while
sibling actions from a shared prefix are correlated. We study whether native MoE router
traces can guide steering and selection without an external judge or selection-time test
execution. Our analysis shows that routing provides a robust behavioral-role signal;
token-granular readouts and decision-matched comparison sets turn it into effective control. We
therefore introduce \textsc{Risa} (\emph{Routing-Informed Steering and Arbitration}):
within trajectories, routing encourages diverse exploration and controlled convergence
during patch commitment; across separately sampled trajectories, agreement at informative
patch positions selects a final candidate. We evaluate on SWE-bench Verified using
open-weight sparse MoE agents across scales and reasoning-effort settings.
\textsc{Risa}'s routing arbitration raises the macro-average resolved rate from $44.9\%$
under uniform sampling to $48.2\%$ on the \texttt{gpt-oss} family, matching text consensus
without answer-string matching, and it transfers to \texttt{Qwen3.6}, where it improves on
uniform choice and matches text consensus on the full $500$-task benchmark.

}

\begin{document}
\maketitle

\begingroup
\renewcommand{\thefootnote}{}
\footnotetext{* Contributed equally.}
\footnotetext{$\dagger$ Corresponding author.}
\endgroup

\section{Introduction}
Repository-level repair unfolds as a sequence of decisions. During one attempt, an agent may inspect
a failing test, search for a symbol, execute a diagnostic command, revise a file, and repeat
this cycle before producing a diff. Giving the agent more inference compute therefore creates
two nested coordination problems. At each tool step, many actions sampled from the same
prefix compete for execution; after several complete runs, multiple non-canonical patches
compete for submission. Although both resemble best-of-$N$ selection, their candidates have
very different dependence: siblings inherit the same trajectory state, whereas complete
attempts may reach a patch through different repository paths.

Existing systems coordinate these choices with execution feedback, surface comparison,
trained verifiers, or judge models
\citep{zhu2025atts,aggarwal2025dars,antoniades2025swesearch}. Sparse mixture-of-experts
(MoE) models expose a complementary signal already produced during inference: for every
token and layer, the router records the selected experts and their weights. Because expert
identities are shared across tokens and actions, this sparse trace can place textually
different candidates in a common coordinate system. It records where the model allocated
internal computation, not only what text it produced. Prior work studies lexical and
semantic specialization in these traces
\citep{xue2024openmoe,muennighoff2025olmoe,olson2025semanticrouting}, but not how their
meaning changes across the nested choices of a long tool-use trajectory.

\begin{figure*}[t]
\centering
\includegraphics[width=0.315\textwidth]{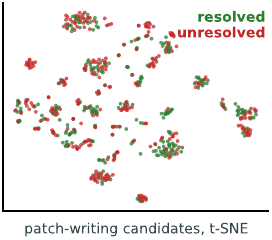}\hfill
\includegraphics[width=0.315\textwidth]{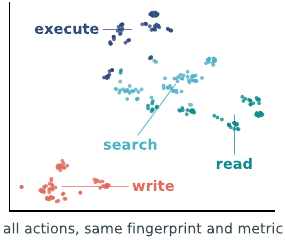}\hfill
\includegraphics[width=0.315\textwidth]{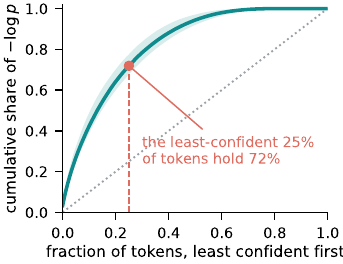}
\caption{Routing supports decision-matched control. \textbf{(a)} Patch-writing fingerprints
overlap across outcomes, motivating relational use. \textbf{(b)} Behavioral roles separate
cleanly. \textbf{(c)} Across $658$ final-step
generations, the least-probable quarter carries $72\%$ of total token surprisal (median and
interquartile range; dotted line: equal contribution).}
\label{fig:teaser}
\end{figure*}

Routing similarity becomes informative relative to the candidates being compared.
Same-prefix siblings share nearly all preceding
evidence, making their agreement partly shaped by a common continuation. Separately sampled
attempts can inspect different files and run different diagnostics, making convergence at
the patch stage a stronger signal. We therefore follow the trace outward through four
questions. Action fingerprints first reveal \emph{what} computation an agent is performing.
Similarity to its executed history then marks computational revisitation. Within long
patches, fine-grained variation concentrates at low-probability \emph{decision tokens}.
Finally, we test how agreement changes as the reference set moves from correlated siblings
to independently developed fixes. Figure~\ref{fig:teaser} previews the role geometry and
decision-token concentration behind this progression.

These findings yield a simple principle: match the routing comparison set to the decision.
We instantiate it as \textsc{Risa} (\emph{Routing-Informed Steering and Arbitration}),
a two-level controller.
Within an attempt, a routing-derived role gate separates exploration from patch writing;
exploration favors actions unlike recent executed history, while a cohort of write
candidates uses guarded peer convergence. Across $K$ attempts, each accumulated patch is
re-encoded once, and the final patch with the highest decision-token agreement across
attempts is selected. Step-level fingerprints come from the original generation pass, and
terminal selection requires no external judge or executing candidate patches solely to
select among them.

Across the six \texttt{gpt-oss} model--effort conditions, \textsc{Risa} improves over Uniform
by $2.3$--$5.7$ points, lifting the macro-average from $44.9\%$ to $48.2\%$; Text reaches
$48.0\%$ and \textsc{Risa-H} $48.3\%$. On the cross-family
full $500$-task \texttt{Qwen3.6-35B-A3B} condition, \textsc{Risa} improves over Uniform
by $3.5$ points ($p<0.001$) and matches Text (exact McNemar $p{=}1.000$), after refitting
only the architecture-dependent role centroids.

Our contributions are as follows:
\begin{itemize}
\item A multiscale map of MoE routing as a behavioral coordinate system, spanning action
roles, trajectory repetition, decision-token detail, and cross-attempt convergence.
\item \textsc{Risa}, a two-level routing-guided controller that turns these findings into
role-conditioned exploration, commitment, and final-patch selection.
\item An evaluation over two model scales, three reasoning-effort settings, and a second
MoE family, in which \textsc{Risa} raises resolved rate over Uniform in every condition while
operating directly on routing traces.
\end{itemize}

\section{Related Work}

\paragraph{Interpreting MoE routing.}
Sparse MoEs expose expert identities and router weights at every token
\citep{shazeer2017moe,fedus2022switch}. Analyses find a strong lexical component
\citep{xue2024openmoe}, expert specialization \citep{muennighoff2025olmoe}, and
same-token semantic sensitivity \citep{olson2025semanticrouting}. These studies establish
token- and corpus-level routing structure. We extend the analysis to what aggregation
preserves over heterogeneous actions, long spans, and different comparison sets.
Routing-agreement decoding (RAD) demonstrates endpoint selection from fixed,
token-aligned routing anchors \citep{chen2026rad}. Accumulated software patches have no
canonical anchor, and long-horizon agents expose a second, same-prefix choice scale. We
therefore localize final patches by teacher-forced token probability and reuse the coordinate
against recent history, role-matched peers, and separately sampled attempts.

\paragraph{Internal signals for test-time control.}
Hidden states can reveal whether a reasoning model's answer or intermediate step is
likely to be correct \citep{zhang2025phsv}, and ReProbe turns them into a trained
step verifier \citep{ni2026reprobe}. Chen et al. instead use sparse-neuron agreement for
label-free best-of-$N$ selection and early pruning \citep{nad2026}. These works use internal states to rank candidate quality. We complement them by mapping
what sparse routing retains at action, trajectory, and attempt scales, then translating its
role- and granularity-specific structure into coordination rules.

\paragraph{Test-time scaling and verification.}
Self-consistency aggregates repeated answer strings \citep{wang2022selfconsistency};
soft and universal variants compare likelihoods or use a separate language model to
select free-form outputs \citep{wang2024softsc,chen2023universal}. Compute-optimal and
confidence-guided scaling further stress the selection signal
\citep{snell2025scaling,fu2025deepconf}. Our contribution is a native-routing
coordination mechanism for repeated samples when long-horizon agents produce
heterogeneous trajectories and non-canonical patches.

\paragraph{Agentic test-time scaling and software repair.}
SWE-bench Verified evaluates repository-level repair against hidden tests
\citep{jimenez2024swebench,chowdhury2024swebenchverified}. Existing agents expose
repository tools or decompose localization and repair \citep{yang2024sweagent,xia2024agentless},
while test-time scaling changes search, feedback, and trajectory reuse
\citep{zhu2025atts,aggarwal2025dars,ding2026swereplay,antoniades2025swesearch}.
Other systems summarize rollouts, train trajectory verifiers, or combine tool entropy with
tests \citep{kim2026agenticcoding,pan2025swegym,mao2026egss}. We complement these
mechanisms with MoE-routing comparisons over recent history, same-prefix candidates, and
separately sampled attempts. Terminal selection uses one same-model prefill per patch, but
no trajectory summary, judge model, or additional candidate-patch execution for selection.

\section{What Routing Reveals Across Agent Decisions}
\label{sec:pilot}

To turn this native MoE trace into control, we first need to understand what survives at
each unit of an agent run. An action may be a short search command or a long patch-writing
invocation; an attempt contains many such actions; and a task yields several independently
sampled attempts. We ask four questions in sequence: what behavior survives aggregation
over an action, whether an action repeats the trajectory's recent computation, where a patch
carries its fine-grained routing variation, and which candidates may meaningfully agree.
The answers become the role gate, history-based exploration rule, decision-token readout,
and granularity-aware commitment rules.

\subsection{From Router Rows to Comparable Fingerprints}
\label{sec:fingerprint}

To compare routing at these scales, we first need one common object. Our agent reads,
searches, edits, and tests before submitting a patch. At each step the model proposes $n$
sibling actions from the same prefix and commits to one. We call the
resulting sequence of executed actions and its accumulated final diff an \emph{attempt};
separately, $K$ attempts produce the final patch candidates. In our configuration,
$n{=}16$ candidate generations are sampled per step and $K{=}4$ attempts per task. The
first choice set shares an immediate prefix, whereas the second can follow different
repository paths. This distinction will determine what agreement means.

A raw router trace contains one sparse expert-weight vector for every token and layer, and
candidate spans have different lengths. To make them comparable, we integrate the gate mass
over a chosen span $s$ and normalize it into a layer-by-expert histogram. Let
$w_{t,\ell,e}$ be the weight assigned by token $t\in s$ to expert $e$ in layer $\ell$, with
zero for an expert that is not selected. The resulting \emph{routing fingerprint} is
\begin{equation}
h(s)_{\ell,e} \;=\; \frac{\sum_{t\in s} w_{t,\ell,e}}{\sum_{t\in s}\sum_{\ell',e'} w_{t,\ell',e'}}.
\label{eq:fingerprint}
\end{equation}
Thus $h(s)$ records where the MoE allocated computation while removing span length. Since
these fingerprints are nonnegative sparse profiles, we compare them with weighted Jaccard,
\begin{equation}
\mathrm{WJ}(u,v) \;=\; \frac{\sum_i \min(u_i, v_i)}{\sum_i \max(u_i, v_i)},
\label{eq:wj}
\end{equation}
where $i$ indexes layer--expert cells. The score is one for identical routing mass and
zero for disjoint mass. Action fingerprints are
read from the original generation pass; accumulated final patches are re-encoded once in
Section~\ref{sec:final} so that the fingerprint represents the submitted artifact itself.

\subsection{Insight 1: Routing Strongly Encodes Behavioral Role}
\label{sec:insight1}

We first ask what survives action-level aggregation. Figure~\ref{fig:teaser}a--b shows a
clear geometry: writes from resolving and non-resolving branches occupy the same broad
representation, while
inspection, search, execution and patch writing separate cleanly. Using parsed tool
calls to label the three roles that the scaffold's commands actually realize---inspect or
execute, run tests, and write---we fit one routing centroid per role within each model
configuration and assign the centroid with highest cosine similarity. On task-disjoint
splits of $78{,}535$
actions, three-way holdout accuracy is $0.940$ against a $0.746$ majority-class baseline;
write-versus-rest recall is $0.93$ at precision $1.00$. For control, the two non-write roles
are collapsed into an exploratory class, while write forms the commitment class. This distinction is operationally important because one sibling set can
mix actions with different purposes: an ordinary read should not become preferable merely
because reads are common when another candidate is ready to write a patch. Routing therefore
supplies a high-precision behavioral gate that first makes candidates commensurable, then
determines which comparison rule to invoke.
Appendix~\ref{app:span-role-gate} gives the centroid data, holdout protocol, and
write-cohort activation rule; Appendix~\ref{app:role-gate} separately audits routing
against exact parsed-tool gating.

\begin{figure*}[!t]
\centering
\includegraphics[width=0.32\textwidth]{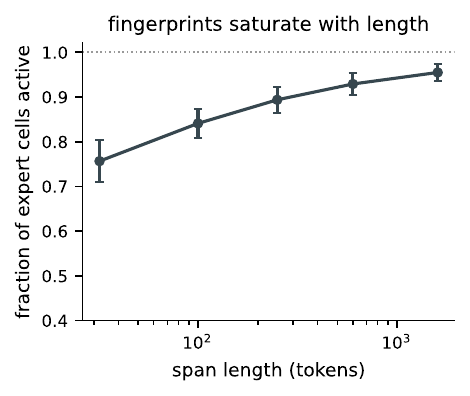}\hfill
\includegraphics[width=0.32\textwidth]{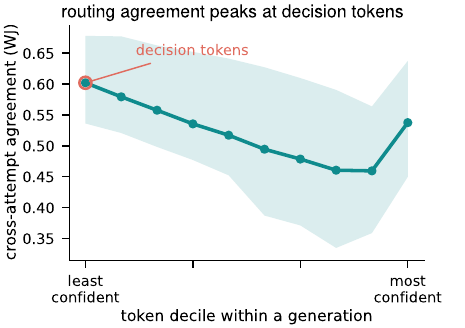}\hfill
\includegraphics[width=0.32\textwidth]{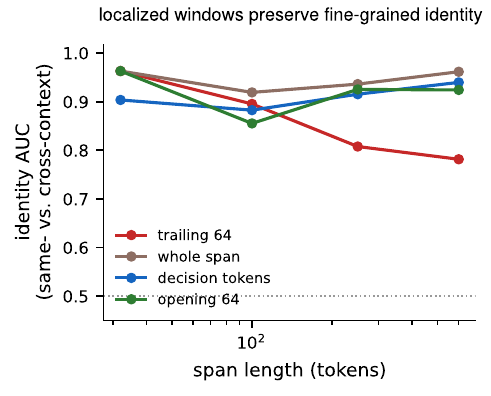}
\caption{Saturation, decision-token localization, and window choice. Left: layer--expert occupancy
(the fraction of cells receiving nonzero mass) grows with span length. Middle: splitting
each generation into ten deciles by next-token probability, routing agreement between
separately sampled attempts is highest in the lowest-probability decile and falls as
probability rises ($1{,}004$ attempt pairs; mean and interquartile range); the final
decile rebounds, consistent with highly predictable tokens being routed alike. Right: the
context-identity AUC (same generation context versus different contexts) decays with length;
whole-span routing retains coarse context, while opening and decision-token windows
preserve finer separation.}
\label{fig:saturation}
\end{figure*}

\subsection{Insight 2: Routing Complements Surface Similarity}
\label{sec:insight2}

We next ask what progress information routing contributes beyond surface form. Once an attempt
has history, each executed action receives a repetition score: its mean WJ similarity to
the three closest fingerprints among up to the previous $64$ actions. Averaging these step scores
within an attempt strongly tracks stagnation: resolved rate falls from $33.8\%$ in the
most-different quintile to $5.4\%$ in the most-similar ($n{=}1{,}021$). Because the
comparison is made in routing space, differently phrased commands can still count as a
revisit when they allocate computation similarly. This trajectory-level signal motivates
disagreement with recent routing history during exploration. Appendix~\ref{app:novelty-shape},
with Figure~\ref{fig:novelty}, reports the quintile construction and contrasts the
within-step and trajectory-level relationships.

At the patch scale, changed-line overlap and routing cover different cases. We score each
diff by mean Jaccard overlap over its added and deleted file--line pairs, after removing
trailing whitespace. This surface score ties at the top on $16.5\%$--$47.4\%$ of tasks
across conditions, whereas routing still orders the candidates. Surface overlap is strongest
when edits match explicitly; routing provides a complementary axis, motivating the hybrid
selector in Table~\ref{tab:main}. Appendix~\ref{app:surface-scores} defines the
changed-line and file-set scores and reports their pairwise AUC and top-tie rates;
Appendix~\ref{app:terminal-ablations}, especially Table~\ref{tab:baselines}, compares
additional surface-centrality controls.

\subsection{Insight 3: Decision Tokens Preserve Fine-Grained Routing Information}
\label{sec:insight3}

We then ask where in a long candidate the router should be read. As a span grows,
layer--expert occupancy (the fraction of cells with nonzero mass) rises from $0.75$ below
64 tokens to $0.96$ beyond a thousand, so a whole-span fingerprint increasingly mixes many
expert pathways. We quantify retained context with a \emph{context-identity AUC}: the AUC for using WJ to
distinguish fingerprint pairs produced under the same candidate-generation context from
pairs produced under different contexts, where a context is the task-and-prefix
state from which a candidate is sampled. On long outputs, whole-span fingerprints reach
$0.88$, decision-token fingerprints $0.93$, and opening-window fingerprints $0.98$; a
trailing window falls to $0.58$. Opening tokens preserve the originating context most
strongly, but they precede many patch-specific choices. We therefore need a localized
window that also coincides with the model's decisions. Appendix~\ref{app:saturation}
reports the full span-length saturation and context-identity controls.

The reason granularity matters is simple: a long patch contains many predictable tokens
from diff syntax, formatting, and locally routine code. If all positions contribute equally,
these common pathways can dominate the fingerprint even when the repair-defining choices
differ. We therefore seek a compact set of positions where the model had to discriminate
among plausible continuations, without assuming a task-specific delimiter.

Token probability provides that localization. Across $658$ final-step generations, the
least-probable quarter carries a median $72\%$ (interquartile range $68$--$78$) of total
token surprisal
$\sum_t-\log p_t$, rather than the $25\%$ expected under equal contribution
(Figure~\ref{fig:teaser}c). We call these positions \emph{decision tokens}: an operational
window that localizes routing variation without requiring a task-specific semantic label for
each token. This concentration defines a compact, generation-agnostic window; matched readouts below test whether it preserves
useful routing variation. For an accumulated final patch, analogous probabilities come from
one teacher-forced re-encoding pass (Section~\ref{sec:final}). Mean cross-attempt routing
agreement is $0.65$ at decision tokens and $0.56$ elsewhere; it generally falls with
confidence before a final-decile rebound (Figure~\ref{fig:saturation}, middle).

Outcome variation also becomes visible at this scale. On write steps containing both
outcomes, we score each candidate by mean WJ agreement with its siblings. Decision-token
routing reaches AUC $0.69$ on shorter candidates, above all other readouts
(Table~\ref{tab:correctness}). Its advantage on compact branch decisions motivates the same
localized readout across independent completed attempts.

\begin{table}[t]
\centering
\small
\begin{tabular}{lcc}
\toprule
 & \multicolumn{2}{c}{candidate length} \\
\cmidrule(lr){2-3}
fingerprint read at & shorter half & longer half \\
\midrule
whole span            & .59 & .52 \\
opening 64 tokens     & .49 & .51 \\
trailing 64 tokens    & .51 & .52 \\
decision tokens       & \textbf{.69} & .52 \\
\bottomrule
\end{tabular}
\caption{Pairwise AUC for separating resolving from non-resolving sibling branches across
3{,}660 pairs from 423 mixed-outcome steps, split at the median candidate length. Decision
tokens give the strongest readout, with their clearest advantage on shorter candidates.}
\label{tab:correctness}
\end{table}

Using each attempt's mean WJ agreement with the other attempts as its score, decision
tokens provide the strongest attempt-ranking readout. In groups containing both outcomes,
the pairwise AUC for ranking a resolving attempt above a non-resolving one is $.657$,
compared with $.639$ for the whole span and $.616$--$.634$ for random, uniformly spaced,
router-entropy, and highest-probability controls. Appendix~\ref{app:window-controls}
enumerates this full control battery, while Appendix~\ref{app:threshold-sensitivity} and
Table~\ref{tab:sens} report window-fraction and neighborhood sensitivity. Fixed delimiters localize routing for
canonical answers \citep{chen2026rad}; low token probability serves as the corresponding
anchor for free-form patches.

\subsection{Insight 4: Granularity Determines How Agreement Should Be Used}
\label{sec:insight4}

\begin{figure*}[t]
\centering
\includegraphics[width=0.32\textwidth]{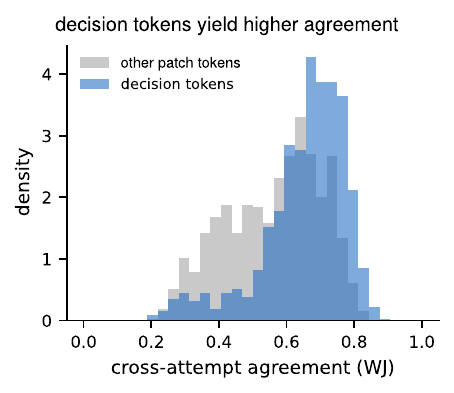}\hfill
\includegraphics[width=0.32\textwidth]{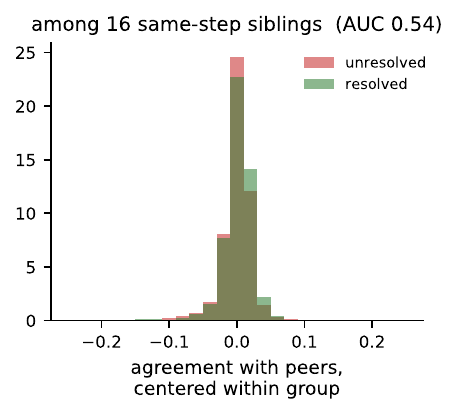}\hfill
\includegraphics[width=0.32\textwidth]{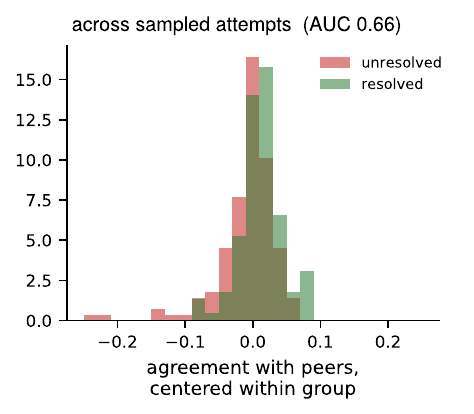}
\caption{Why the reference set matters. Left: decision tokens yield higher cross-attempt
agreement than other positions. Middle: same-prefix outcome distributions overlap, motivating
history-relative novelty. Right: across separately sampled attempts, resolving patches receive higher mean agreement
with their attempt group.}
\label{fig:granularity}
\end{figure*}

The final question is whose agreement is meaningful. Same-prefix siblings share most of
their evidence, so their outcome-conditioned agreement distributions overlap
(Figure~\ref{fig:granularity}, middle). Exploration therefore compares candidates with recent
executed history (Section~\ref{sec:insight2}). Peer support enters only after the
role gate forms a write cohort, where a pilot-fixed composite combines local support with
routing-dispersion guards (Section~\ref{sec:steering}). Appendix~\ref{app:write-rule}
specifies the standardized score, fixed coefficients, and guard terms, while
Appendix~\ref{app:write-score}, especially Table~\ref{tab:writecomp}, ablates the
individual components.

Across separately sampled attempts, agreement becomes a direct ranking signal. The
trajectories follow different repository paths and share no immediate prefix; mean
agreement between their decision-token fingerprints ranks resolving attempts above
non-resolving ones with AUC $0.66$ in mixed-outcome groups. Together, these regimes yield
the operational rule in the title: disagree with
recent computation to explore; use agreement only after role gating, and across
trajectories to commit.

\subsection{From Observations to Design}

\begin{figure*}[t]
\centering
\includegraphics[width=0.96\textwidth]{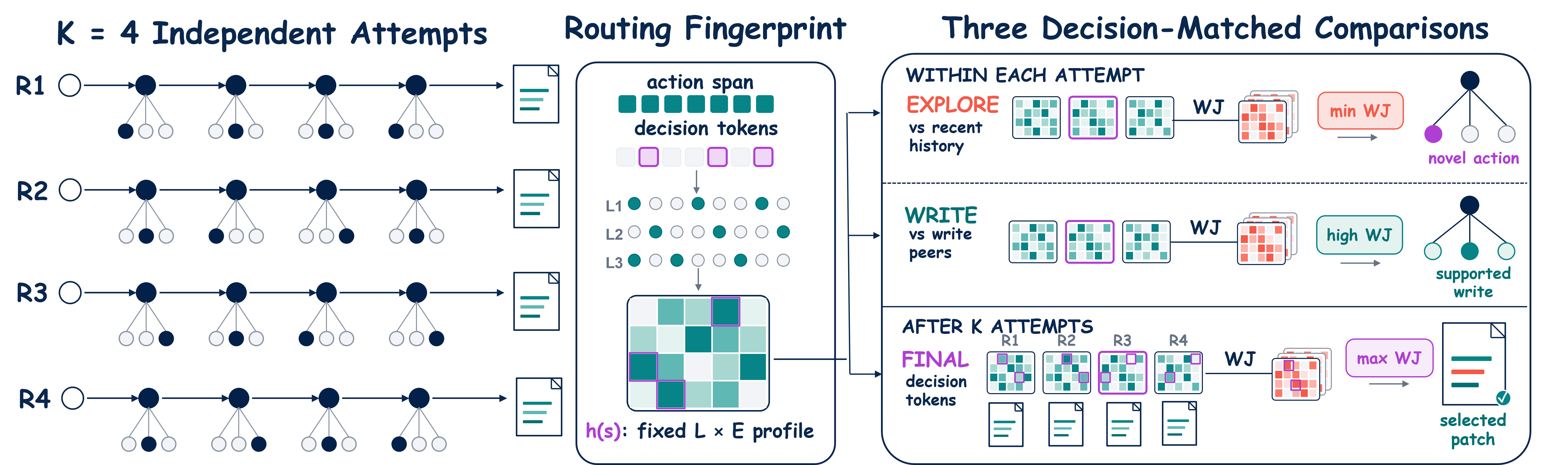}
\caption{\textsc{Risa}: one MoE routing representation, three decision-matched comparisons. Exploratory
actions are compared with recent executed history; patch-writing actions with their
role-gated peers; and accumulated final patches with separately sampled attempts at
decision tokens.}
\label{fig:pipeline}
\end{figure*}

MoE routing is most useful here as a shared computational coordinate: the same
layer--expert axes describe textually different actions and patches. Routing similarity derives its decision meaning from the chosen reference set. A high value can indicate stagnation against recent history,
useful support among role-matched writes, or convergence across independent attempts. The
controller must consequently change both the comparison set and the direction in which
similarity is optimized. Against recent executed history, similarity measures computational
revisitation; within a role-matched write cohort, guarded centrality measures convergent
commitment; across separately sampled attempts, decision-token centrality measures cross-path
convergence. The fingerprint supplies the coordinate, while the comparison set supplies the
semantics. Figure~\ref{fig:pipeline} turns these three relations into \textsc{Risa}.

\section{Method: \textsc{Risa}}
\label{sec:method}

Inside each attempt, \textsc{Risa} repeats four steps: sample sibling generations, isolate and fingerprint their
action spans, use the routing role gate to choose a comparison rule, and execute the selected
action while updating history. After $K$ complete attempts, \textsc{Risa} re-encodes the accumulated
diffs and applies decision-token agreement once to choose the submission. The model weights
remain fixed online, and the same MoE telemetry is reused at each scale by changing its
reference set rather than training a separate scorer for every decision.

Step-level fingerprints come from the original generation. Three role centroids
(inspect/execute, test, write) are fit per model configuration from parsed
tool-call labels; at control time, each candidate is assigned to its highest-cosine centroid
and the two non-write roles are grouped as exploratory. Architecture-specific centroids are
fit from task-disjoint action labels, while the write-score coefficients are fixed on a
separate pilot. Both are frozen before evaluation; role definitions, decision logic, and
coefficients are shared across tasks and effort settings. The behavioral gate assigns candidates to the comparison rule suited to their role; the
relation-specific score then ranks them. Parsed calls provide convenient supervision, while
routing provides the portable runtime signal: $97.3\%$ of writes here are shell commands
inside a generic execution tool, whose runtime parsing requires a scaffold-specific command
taxonomy. Reading routing keeps both levels on one signal and agrees with the exact
parsed role on $96.8\%$ of actions. Appendix~\ref{app:parser-taxonomy} details the
command taxonomy required by the parsed gate, and Appendix~\ref{app:role-substitution}
reports the controlled gate-substitution test.

\subsection{Steering a Single Attempt}
\label{sec:steering}

At step $t$, the model samples $n{=}16$ candidate generations $\mathcal C_t$ in one
batched call. A generation may contain reasoning followed by a proposed tool invocation;
we fingerprint only the \emph{action span} that serializes the invocation, located from the
model's action-channel markers, so the comparison describes the proposed action rather than
its preceding reasoning. We write $h(c)$ for this fingerprint. The history $\mathcal H_t$
stores the last $W{=}64$ selected and executed action fingerprints.
Appendix~\ref{app:span-role-gate} specifies the span fallbacks: the trailing $64$
generated tokens are used when action markers are missing, and accumulation becomes
unweighted when gate weights are unavailable.

The role prediction is collapsed into write versus exploratory. A cohort of at
least two writes activates the commitment rule, because peer agreement is only defined when
there is another role-matched proposal to support it; otherwise all candidates use the
history-disagreement rule. The selected action is then executed and its fingerprint appended
to $\mathcal H_{t+1}$.

\paragraph{Exploration: disagree with recent history.}
When the write rule is inactive and history is nonempty, set
$m_t=\min(3,|\mathcal H_t|)$ and score
\begin{equation}
S_{\mathrm{explore}}(c)
=-\frac{1}{m_t}\sum_{u\in\mathcal N_{m_t}(c)}
\mathrm{WJ}\bigl(h(c),u\bigr),
\label{eq:explore}
\end{equation}
where $\mathcal N_{m_t}(c)$ contains the $m_t$ recent-history entries most similar to $c$;
the maximum is selected. Closest matches expose repeated computation even when the rest of
the history is unrelated. With no history, the controller uses a deterministic cold-start
fallback. Appendix~\ref{app:cold-start} specifies that it uses the sibling routing medoid
at the first eligible step and falls back to the first exchangeable sample only when no
fingerprint is available. Thereafter, Equation~\ref{eq:explore} measures novelty relative to the
attempt's own executed computation, not to same-prefix siblings.

\paragraph{Patch writing: commit toward peer agreement.}
Once several candidates propose writes, continuing to maximize novelty can delay a patch
that multiple branches have independently supported. At the same time, blindly taking the
global medoid may favor an uninformative routing concentration. We therefore combine local
support and cohort-wide similarity with two dispersion guards. Let $\mathcal C_t^{w}$ be
the role-gated write cohort. For $|\mathcal C_t^{w}|\geq 2$, local peer support is the
fraction of other write candidates
within the fixed routing-similarity threshold $\tau{=}0.65$:
\begin{equation}
M(c)=\frac{1}{|\mathcal C_t^{w}|-1}
\sum_{\substack{c'\in\mathcal C_t^{w}\\c'\neq c}}
\mathbf 1\!\left[\mathrm{WJ}\bigl(h(c),h(c')\bigr)\geq\tau\right].
\label{eq:write-support}
\end{equation}
The score also uses mean peer similarity
$\bar S(c)=(|\mathcal C_t^w|-1)^{-1}
\sum_{c'\in\mathcal C_t^w,\,c'\neq c}\mathrm{WJ}(h(c),h(c'))$,
fingerprint entropy $H(c)=-\sum_i h_i(c)\log h_i(c)$, and peak mass
$P(c)=\max_i h_i(c)$, where $i$ indexes layer--expert cells. Each statistic is standardized
within the current write cohort as
$z(x)=(x-\mu_x)/\sigma_x$ (a zero-variance statistic contributes zero), and the
controller selects the maximizer of
\begin{equation}
S_{\mathrm{write}}(c)=z\bigl(H(c)\bigr)+\tfrac{3}{2}\,z\bigl(M(c)\bigr)
+z\bigl(\bar S(c)\bigr)-z\bigl(P(c)\bigr).
\label{eq:write-score}
\end{equation}
Here $M$ captures local support, $\bar S$ cohort centrality, and $H$ and $P$ guard against
a concentrated routing spike. The coefficients were fixed once on a disjoint pilot and are
shared across tasks, models, and effort settings; commitment therefore rewards broad,
role-consistent convergence rather than unconditional sibling majority.

\subsection{Arbitrating Among Multiple Attempts}
\label{sec:final}

Each task runs $K{=}4$ attempts; let $\mathcal A$ index those ending with a nonempty
accumulated final diff. Because later edits can overwrite earlier ones, no single generated
action span necessarily represents the submitted artifact. We therefore fingerprint the
artifact directly. For $k\in\mathcal A$, the task statement concatenated with final diff
$p_k$ is re-encoded in one teacher-forced pass, yielding a routing row and next-token
probability for every patch token. This pass consumes the fixed patch without sampling new
text, so all candidates are measured under the same task-statement context. Let $D_k$ be its
least-probable quarter; aggregating
Equation~\ref{eq:fingerprint} over $D_k$ gives the decision-token fingerprint
$h_{\mathrm{dec}}(p_k)$. All attempts use the same task-statement prefix.

When $|\mathcal A|\geq2$, the submitted patch maximizes mean agreement with the other
separately sampled attempts:
\begin{equation}
\hat{k} \;=\; \arg\max_{k\in\mathcal A}\;
\frac{1}{|\mathcal A|-1}\sum_{\substack{j\in\mathcal A\\j\neq k}}
\mathrm{WJ}\bigl(h_{\mathrm{dec}}(p_k), h_{\mathrm{dec}}(p_j)\bigr).
\label{eq:final}
\end{equation}
This surface-overlap-free rule selects the decision-token routing medoid of the available
patches. A sole patch is submitted directly and an empty pool is unresolved. Selection adds
one teacher-forced prefill per available patch but no candidate generation or execution;
Appendix~\ref{app:method-cost} quantifies this terminal overhead alongside the batched
step-generation cost.

\section{Experiments}

\paragraph{Setup.}
We evaluate on SWE-bench Verified with \texttt{gpt-oss-20b} and
\texttt{gpt-oss-120b} at their native low, medium, and high reasoning-effort settings.
Each step considers $n{=}16$ siblings and each task runs $K{=}4$ attempts. Official grading
yields task-level resolved rate. In Table~\ref{tab:main}, \emph{Yield} is the fraction of
completed attempts ending with a nonempty final diff and \emph{Steps} the mean executed
actions per attempt. Selector rates use 496--498 tasks per condition with at least two
graded patches; every terminal rule receives the same \textsc{Risa}-steered pool.
Appendix~\ref{app:grading} details the shared environmental exclusions and selector
denominator, while Appendix~\ref{app:setting-selection} records how the final budgets and
routing-readout settings were chosen.

\paragraph{Terminal arbitration.}
All rules act on the same \textsc{Risa}-steered pools. \emph{Uniform} is expected random
choice; \emph{Text} maximizes mean Jaccard over added/deleted file--line pairs after trimming
trailing whitespace. \textsc{Risa} uses Equation~\ref{eq:final}; \textsc{Risa-H} is Text-first
with routing tie-breaking. \emph{Oracle} is the attempt-pool success union.
Appendix~\ref{app:tie-breaking} specifies the score tolerances, missing-fingerprint
behavior, and fixed-index tie rules used by all selectors.

\begin{table*}[t]
\centering
\scriptsize
\setlength{\tabcolsep}{3.2pt}
\renewcommand{\arraystretch}{1.18}
\begin{tabular}{llccccc>{\columncolor{highlight}}ccc}
\toprule
& & & \multicolumn{2}{c}{\textbf{Attempt pool}} &
\multicolumn{4}{c}{\textbf{Terminal arbitration}} &
\multicolumn{1}{c}{\textbf{Upper bound}} \\
\cmidrule(lr){4-5}\cmidrule(lr){6-9}\cmidrule(lr){10-10}
\textbf{Model} & \textbf{Effort} & \textbf{Eligible} & \textbf{Yield} & \textbf{Steps}
& \textbf{Uniform} & \textbf{Text}
& \textbf{\textsc{Risa}} & \textbf{\textsc{Risa-H}} & \textbf{Oracle} \\
\midrule
\texttt{gpt-oss-20b}  & low    & 496 & 79.1 & 9.5 & 31.8 & \secondcell{34.5}{2.7}
& 34.1\,\gain{2.3} & \bestcell{34.7}{2.9} & 48.4 \\
                       & medium & 498 & 94.8 & 26.5 & 45.7 & \secondcell{47.8}{2.1}
& \bestcell{48.2}{2.5} & 47.6\,\gain{1.9} & 63.3 \\
                       & high   & 498 & 80.0 & 43.1 & 45.3 & 49.2\,\gain{3.9}
& \bestcell{51.0}{5.7} & \secondcell{50.0}{4.7} & 62.2 \\
\midrule
\texttt{gpt-oss-120b} & low    & 497 & 93.0 & 14.8 & 38.7 & 40.0\,\gain{1.3}
& \bestcell{41.2}{2.5} & \secondcell{40.4}{1.7} & 53.3 \\
                       & medium & 497 & 99.5 & 28.8 & 50.6 & \secondcell{54.5}{3.9}
& 53.5\,\gain{2.9} & \bestcell{54.9}{4.3} & 66.6 \\
                       & high   & 498 & 94.6 & 41.9 & 57.2 & \secondcell{62.0}{4.8}
& 61.2\,\gain{4.0} & \bestcell{62.4}{5.2} & 71.5 \\
\midrule
\rowcolor{gray!7}
\multicolumn{2}{l}{\textsc{Macro Avg.} (\texttt{gpt-oss})} & --- & 90.2 & 27.4
& 44.9 & 48.0\,\gain{3.1} & \secondcell{48.2}{3.3}
& \bestcell{48.3}{3.5} & 60.9 \\
\midrule
\texttt{Qwen3.6-35B-A3B} & default & 498 & 73.5 & 47.5 & 41.7 & 45.0\,\gain{3.3}
& \secondcell{45.2}{3.5} & \bestcell{45.6}{3.9} & 50.6 \\
\bottomrule
\end{tabular}
\caption{Final-patch arbitration on SWE-bench Verified (\% resolved) over common
\textsc{Risa}-steered $K{=}4$ pools. Rates condition on at least two graded patches. The full
$500$-task \texttt{gpt-oss} grid yields $496$--$498$ eligible tasks per condition; the full $500$-task
\texttt{Qwen3.6-35B-A3B} condition has $498$. \textsc{Risa-H} uses Text-first routing
tie-breaking; Uniform is expected random choice and Oracle the success union. Parentheses
in deployable columns show gains over Uniform; bold shading and underlining mark the best and
second-best deployable estimates.}
\label{tab:main}
\end{table*}

\subsection{Main Results}

Across the six \texttt{gpt-oss} conditions, \textsc{Risa} gains $2.3$--$5.7$ points over Uniform,
lifting the macro-average from $44.9\%$ to $48.2\%$; the largest gains occur at high
reasoning effort. \textsc{Risa} is best in three conditions and \textsc{Risa-H} in the other three, with
macro-averages of $48.2\%$ and $48.3\%$, respectively, versus $48.0\%$ for Text. The $60.9\%$ Oracle confirms substantial complementary coverage across the four-attempt
pools. Appendix~\ref{app:statistics} reports the task-paired bootstrap intervals and
McNemar tests; Appendix~\ref{app:terminal-ablations}, especially
Table~\ref{tab:baselines}, compares alternative surface and routing centrality rules.

The same rules transfer to \texttt{Qwen3.6-35B-A3B}; only its architecture-dependent role
centroids are refit from task-disjoint action labels. On the $498$ eligible tasks, \textsc{Risa}
resolves $45.2\%$, versus $41.7\%$ for Uniform and $45.0\%$ for Text; \textsc{Risa-H}
reaches $45.6\%$. Both \textsc{Risa} and Text significantly outperform Uniform
($p<0.001$). \textsc{Risa} matches Text, with 10 routing-only wins and 9 text-only wins
(exact McNemar $p{=}1.000$). This cross-family condition is
reported separately from the \texttt{gpt-oss} macro-average in
Appendix~\ref{app:qwen}, Table~\ref{tab:qwen}.

\subsection{Analysis}

\paragraph{The fixed-pool comparison isolates arbitration.}
Every selector in Table~\ref{tab:main} receives the same \textsc{Risa}-steered attempts, so
the differences isolate terminal arbitration from candidate generation and step budget.
Routing improves over Uniform in every reported condition, and the gains persist as Yield
ranges from $79.1\%$ to $99.5\%$, showing robustness across candidate availability.

\paragraph{Routing matches and complements surface consensus.}
\textsc{Risa} is best in three \texttt{gpt-oss} conditions and \textsc{Risa-H} in the other
three; their $48.2\%$ and $48.3\%$ macro-averages edge Text's $48.0\%$, supporting
complementarity on the \texttt{gpt-oss} grid. On the full Qwen condition, \textsc{Risa}
matches Text ($45.2\%$ versus $45.0\%$; exact McNemar $p{=}1.000$). The cross-family
result shows that routing-only arbitration maintains text-consensus performance while
operating directly on internal routing traces.

\paragraph{Decision tokens provide the strongest terminal readout.}
As Section~\ref{sec:insight3} shows, the least-probable quarter ranks mixed-outcome attempts
better than the whole span and matched controls, motivating the terminal readout. Applied to
identical attempt pools, this localized representation yields gains in every reported
condition, and the same decision-token rule transfers to Qwen as a significant $3.5$-point
gain over Uniform. The token-level analysis thus connects directly to final-patch arbitration.

\paragraph{Supporting diagnostics.}
Table~\ref{tab:main} deliberately holds generation fixed. Appendix~\ref{app:steering-coverage}
isolates steering through patch-yield coverage on the empirically hard $80$-instance set,
while Appendix~\ref{app:pipeline}, Table~\ref{tab:costmatched}, decomposes steering and
terminal arbitration on a fixed $200$-task pool. On the hard set, steering
raises submittable-patch yield from $79\%$ to $94\%$; on a fixed $200$-task \texttt{20b}
subset, the full pipeline reaches $50.5\%$ versus $45.4\%$ for unguided generation with
Uniform selection. Routing also improves selection on pools generated without steering,
showing that steering expands candidate availability while arbitration contributes independently.

\section{Conclusion}

Sparse-MoE routing gives software agents a reference-dependent computational coordinate
for coordinating nested decisions. This view yields \textsc{Risa}: novelty against
recent history guides exploration, guarded peer support guides writing, and decision-token
agreement across independently developed patches guides final selection. Across six
\texttt{gpt-oss} conditions, routing arbitration gains $2.3$--$5.7$ points over Uniform, reaching a
$48.2\%$ macro-average versus $48.0\%$ for Text. On the full Qwen benchmark, it reaches
$45.2\%$ versus $41.7\%$ for Uniform and $45.0\%$ for Text, transferring as a significant
gain over uniform selection while matching text consensus. Together, these results establish
routing as a practical coordination signal that transfers across MoE families.

\section{Limitations}

\textsc{Risa}'s current instantiation assumes accessible sparse-MoE routing and repeated
trajectories, making it naturally suited to white-box MoE agents. The broader principle is to
construct a shared internal coordinate, align it with behavioral roles, and choose reference
sets that match the decision. Extending this principle to dense or closed models will require
alternative readouts, while other domains will need role definitions suited to their action
spaces. Combining routing coordination with semantic or execution-based evidence is a
promising direction for handling rare but valuable outlier repairs.

\clearpage
\bibliographystyle{plainnat}
\bibliography{references}

\appendix
\setcounter{secnumdepth}{3}
\section{Implementation Details}
\label{app:implementation}

\subsection{Models and Serving}
\label{app:models-serving}
The completed main grid uses the open-weight sparse mixture-of-experts models
\texttt{gpt-oss-20b} (24 transformer blocks, 32 experts per MoE layer, top-4 routing,
$\sim$3.6B active parameters) and \texttt{gpt-oss-120b} (36 blocks, 128 experts, top-4
routing, $\sim$5.1B active parameters), served with vLLM 0.15.1 through the raw
\texttt{/v1/completions} endpoint. The cross-family condition uses
\texttt{Qwen3.6-35B-A3B} (40 blocks, 256 experts, top-8 routing, $\sim$3B active
parameters) on the same stack, and is reported separately from the \texttt{gpt-oss} grid.
For the \texttt{gpt-oss} grid, conversations are rendered in the native Harmony token format
and parsed back into analysis / tool / final channels; tools are declared in the system
message. The serving stack returns, for every generated request,
the per-token expert assignments (\texttt{routed\_experts}, an integer tensor of shape
$[T, L, R]$ for $T$ tokens, $L$ MoE layers, and $R{=}4$ routing slots) and the corresponding
post-softmax gate weights (\texttt{routed\_experts\_weights}, same shape, rows summing
to one). Both are recorded at generation time for every candidate action, together with
next-token probabilities. Step-level steering needs no additional model pass; terminal
selection adds one teacher-forced prefill per available final patch
(Section~\ref{app:terminal-arbitration}).

\subsection{Agent Scaffold}
\label{app:agent-scaffold}
The agent is a single-model tool-use loop over an isolated Docker container per task
(SWE-bench instance image; the model edits the repository at \texttt{/testbed} and the
final \texttt{git diff} is the submitted patch). Available tools are the model's
in-distribution set: \texttt{container.exec}, \texttt{repo\_browser.*} (search, tree,
open file), and \texttt{apply\_patch}. At every step the model proposes $n{=}16$
candidate actions sampled in parallel from the identical prefix; one candidate is
selected by the exploration or write rule
(Sections~\ref{app:exploration-rule} and~\ref{app:write-rule}) and executed.
At an explicit submission or budget exhaustion,
the current \texttt{git diff} is retained as the final patch. Each task runs $K{=}4$
separately sampled trajectories.

\subsection{Routing Fingerprints and \textsc{Risa} Rules}
\label{app:risa-rules}
Throughout, \textsc{Risa} (\emph{Routing-Informed Steering and Arbitration}) denotes
routing-guided steering plus routing-only terminal
arbitration; \textsc{Risa-H} keeps the same steering but uses the Text-first routing tie-breaker.
\subsubsection{Fingerprint}
\label{app:fingerprint}
For a token span $S$ of a candidate, the fingerprint
$h \in \mathbb{R}^{L \times E}$ accumulates gate mass,
\[
h[l,e] \;=\; \sum_{t \in S} \sum_{q=1}^{R} w_{t,l,q}\,\mathbf{1}[\,r_{t,l,q}=e\,],
\]
where $r_{t,l,q}$ is the expert index and $w_{t,l,q}$ its gate weight; $h$ is
$L_1$-normalized. Similarity between fingerprints is the weighted Jaccard
(Ruzicka) score $\mathrm{WJ}(a,b) = \sum_i \min(a_i,b_i) / \sum_i \max(a_i,b_i)$,
where $i$ indexes layer--expert cells.

\subsubsection{Candidate Span and Role Gate}
\label{app:span-role-gate}
A candidate generation may contain
reasoning followed by a proposed tool invocation. Its \emph{action span} is the token
segment that serializes the final invocation, delimited by the model's action-channel marker
tokens rather than decoded-string matching. If the markers cannot be located, the span falls
back to the
trailing 64 generated tokens; if gate weights are unavailable, accumulation is unweighted.
We fit three centroids from actions labeled by parsed tool calls: inspect/execute (repository
browsing, search and other shell commands, $53{,}923$ actions), test (test-runner commands,
$8{,}183$) and write (patch application or an editing shell command, $16{,}429$). Reading and
searching are issued through the same generic execution tool as other commands and are not
separated further, since only the write/non-write boundary gates a rule. At control time, the
role whose centroid has highest cosine similarity to the candidate's
routing fingerprint is predicted; the two non-write roles are collapsed to the exploratory
class. Parsed labels supervise the centroids, while the deployed gate
uses the routing prediction. Since $L\times E$ differs by architecture, centroids are fit
separately within
each model configuration using the same role definition. On task-disjoint 70/30 splits of
$78{,}535$ actions, three-way holdout accuracy is $0.940$ against a $0.746$ majority floor
(the released centroid file carries exactly these three labels);
write-versus-rest recall is $0.93$ at precision $1.00$. At least two predicted writes are
required to activate commitment; otherwise all candidates remain under the exploration
rule.

\subsubsection{Diagnostic Branch Labels}
\label{app:branch-labels}
For branch-level analyses, a write candidate carries
the downstream official resolved/unresolved label of its evaluated branch. A
\emph{mixed-outcome step} contains at least one branch of each label. Pairwise area under
the ROC curve (AUC) treats a resolved--unresolved pair as correctly ordered when the score
ranks the resolving branch higher; the ``resolved'' and ``unresolved'' labels in main-paper
figures refer to these downstream outcomes.

\subsubsection{Exploration Steps: Disagree}
\label{app:exploration-rule}
Each candidate's action fingerprint is
compared with the agent's own recently executed actions. With
$m=\min(3,|\mathcal H_t|)$, we average the $m$ largest similarities and select the
candidate whose average is lowest; using the closest matches prevents one repeated
computation from being masked by unrelated older entries. The history pool contains up to
the last $W{=}64$ executed-action fingerprints, with no role matching. If history is
empty, the sibling routing medoid---the candidate with greatest mean WJ to the other
siblings---is used as a deterministic cold-start fallback, not as an outcome-ranking rule;
if candidate fingerprints are unavailable, the rule samples uniformly.

\subsubsection{Write Steps: Guarded Peer Support}
\label{app:write-rule}
Let $\mathcal C_t^w$ be the
predicted write cohort. If $|\mathcal C_t^w|\geq2$, local peer support is
\[
M(c)=\frac{1}{|\mathcal C_t^w|-1}\sum_{c'\neq c}
\mathbf 1[\mathrm{WJ}(h(c),h(c'))\geq\tau],
\]
with $\tau{=}0.65$. The other terms are
fingerprint entropy $H(c)=-\sum_i h_i(c)\log h_i(c)$, mean peer similarity
$\bar S(c)=(|\mathcal C_t^w|-1)^{-1}\sum_{c'\neq c}\mathrm{WJ}(h(c),h(c'))$, and
largest layer--expert-cell mass $P(c)=\max_i h_i(c)$. Each statistic is standardized over
the current cohort as $z(x)=(x-\mu_x)/\sigma_x$; a zero-variance term contributes zero.
The fixed score is
\[
S_{\mathrm{write}}(c)=z(H(c))+1.5\,z(M(c))+z(\bar S(c))-z(P(c)),
\]
and the maximizer is selected. Here $M$ captures a local agreement neighborhood and
$\bar S$ global centrality, while $H$ and $P$ guard against agreement caused by a highly
concentrated routing spike. The coefficients $(1,1.5,1,-1)$ are constants in the released
configuration: they are not refit per task, model, or effort setting, and no per-run
optimization takes place. Their signs and relative magnitude were fixed once, before the
main campaign, from a pilot study of $98$ mixed-outcome write steps ($650$
resolved--unresolved candidate pairs) in which this combination reached a step-clustered
bootstrap AUC of $0.592$ (95\% confidence interval $[0.512,0.672]$) for
ranking the resolving branch higher;
that pilot pool is disjoint from the tasks used in the ablations reported here. The
pilot-fitted scorer remains fixed throughout evaluation. Fewer than two predicted writes returns the full
candidate set to the exploration rule.

\subsubsection{\textsc{Risa} Arbitration Across Multiple Attempts}
\label{app:terminal-arbitration}
An available final patch is a nonempty accumulated \texttt{git diff}, which may
combine hunks written at several steps, some overwriting earlier edits. We therefore
re-encode each complete final patch once: the task statement
concatenated with the patch is sent as a single prompt with \texttt{max\_tokens}${=}1$
and \texttt{temperature}${=}0$, and the server returns a routing row and a teacher-forced
token probability for every prompt position; the required one-token completion is
discarded. Only positions inside the patch segment are
used; the offset of that segment is the token length of the task statement, measured by
the same request. Positions in the lowest probability quartile of the patch segment (the
\emph{decision tokens}) contribute their gate-weighted routing rows to the attempt
fingerprint. The probabilities are the model's teacher-forced next-token probabilities
before any sampling transform, so they do not depend on the temperature or nucleus
setting used during the rollout, and all $K$ attempts of a task are encoded under an
identical prefix. The submitted attempt has the highest mean WJ agreement with the other
available attempts; a single available patch is submitted directly, and a pool with no
patch is unresolved. Terminal selection costs one prefill plus one discarded token per
available patch; it neither re-executes a patch nor generates an additional patch.
Fractions from 10\% to 50\% form a stable operating region, changing cross-attempt ranking
AUC by at most $0.01$.

\subsection{Diagnostic Pool, Surface Scores, and Tie-Breaking}
\label{app:diagnostics}

\subsubsection{Hard Diagnostic Pool}
\label{app:hard-pool}
The $80$-task pool used for the within-trajectory
policy comparisons is defined by \emph{empirical} difficulty measured from agent behavior.
In a pilot phase each Verified task was attempted six times, giving a solve
count in $\{0,\dots,6\}$; from that ranking we took every task solved once
($55$ tasks), a seeded random $10$ with zero solves, and a seeded random $31$ of
those solved twice, for $96$ tasks, then made a difficulty-stratified $80/16$
split whose $80$-task side is the pool used here (\texttt{seed}${=}42$, both lists
released with the code). Its composition is $36$ \texttt{django}, $13$ \texttt{sympy},
$11$ \texttt{sphinx-doc}, $5$ \texttt{matplotlib}, and nine or fewer each from five
other repositories. Its benchmark-label distribution is $23$
\texttt{<15 min}, $48$ \texttt{15 min--1 hour}, $8$ \texttt{1--4 hours}, $1$
\texttt{>4 hours}. Its behavioral criterion selects tasks solved at most twice in six
attempts, producing a focused diagnostic where uniform sibling choice resolves $15.9\%$.

\subsubsection{Surface Scores}
\label{app:surface-scores}
All text comparisons use one of two sets extracted from a
unified diff. The changed-line set $H(p)$ contains one element per added or deleted
line, namely the pair (target file, line content with trailing whitespace removed);
target files are read from \texttt{+++ b/} headers and the \texttt{---}/\texttt{+++}
header lines themselves are skipped. The touched-file set $F(p)$ contains the targets of
\texttt{+++ b/} and \texttt{diff --git} headers. Both are compared with the Jaccard
index $J(A,B)=|A\cap B|/|A\cup B|$, defined as $0$ when both sets are empty, so two
empty patches are not treated as agreeing. The Text selector scores each attempt by the
mean $J(H(p_k),H(p_j))$ over the other available attempts; the file-level variant
substitutes $F$. ``Surface-overlap ties'' in Insight~2 means that this mean is
identical for the top two attempts to within $10^{-6}$. On resolving--non-resolving
cross-attempt pairs with unequal surface scores, changed-line centrality reaches pairwise
AUC $.64$; over the complete pair set, ties reduce it to $.52$, while routing retains
$.58$. Depending on the model--effort condition, changed-line centrality ties at the top on
$16.5\%$--$47.4\%$ of eligible tasks.

\subsubsection{Tie-Breaking}
\label{app:tie-breaking}
Attempts are ordered by a fixed arm index, and every selector
takes the first attempt whose score exceeds the running maximum by more than $10^{-9}$;
ties therefore resolve to the lowest arm index, which is independent of the patches and
of the routing. An attempt whose fingerprint is missing is scored $-\infty$ and is
chosen only if no attempt has one. \textsc{Risa-H} first computes the Text score;
if the top two attempts are within $10^{-6}$ it re-ranks the tied set by routing
agreement under the same first-index rule, and otherwise returns the Text choice. Within
a trajectory, the step-level rules use \texttt{argmax} over the candidate list in
generation order, so ties there resolve to the lowest candidate index.

\subsection{Grading Protocol}
\label{app:grading}
Patches are graded with the official SWE-bench harness (fail-to-pass and pass-to-pass
tests inside a fresh instance container). We grade every stored final patch regardless
of the agent's exit status. All terminal selectors within a model--effort condition use
the same evaluable task and attempt pool.

The shared main-table eligibility rule yields 496--498 tasks per model--effort condition.
Two Verified instances (\texttt{astropy-8707}, \texttt{astropy-8872}) encounter the same
reproducible image-build incompatibility in their pinned 2019 build stack, including with
unrestricted network access. Patchless attempts remain represented in the Yield column and
count as unresolved in attempt-level analyses, while terminal selectors operate on available
candidate patches. Selector comparisons include tasks with at least two graded patches,
yielding up to two additional eligibility differences per condition. Shared eligibility
preserves the paired comparisons; using all 500 tasks changes absolute rates uniformly by at
most $0.5$ points.

\section{Statistical Tests}
\label{app:statistics}

All selector comparisons are paired: every selector acts on the same $K$ attempts of the
same task, so the unit of analysis is the task. Against Uniform we compute, per task, the
difference between the selector's outcome and the expected outcome of choosing one
available candidate patch uniformly, and bootstrap that difference with $10^4$ resamples;
selectors are compared with each other by McNemar's exact test on the discordant tasks.

The $2{,}984$ pooled rows reuse the same $498$ task IDs across up to six
model--effort conditions rather than providing $2{,}984$ independent observations. Pooled
resampling is therefore clustered on the task: a bootstrap replicate draws tasks with
replacement and keeps all available conditions for each drawn task. The clustered and
unclustered intervals align closely (\textsc{Risa}${-}$Uniform $[+2.46,+4.19]$ clustered
versus $[+2.47,+4.16]$ unclustered), because the quantity being resampled is a
\emph{difference} against the same task's uniform expectation, which is far less
correlated across conditions than the accuracies themselves. Each condition contains
every task at most once, so per-condition intervals need no clustering.

Per condition, \textsc{Risa}${-}$Uniform is $+2.3$, $+2.5$, $+5.7$ points for
\texttt{gpt-oss-20b} at low, medium and high effort and $+2.5$, $+2.9$, $+4.0$ points
for \texttt{gpt-oss-120b}, with unadjusted $p$ from $.042$ to $<.001$. These condition-level
$p$-values provide descriptive arm-wise summaries, while the task-clustered bootstrap gives
the aggregate cross-condition estimate. Text and \textsc{Risa-H} reach $+3.12$ and $+3.46$ points over Uniform pooled,
against $+3.33$ for \textsc{Risa}. \textsc{Risa} versus Text is $+0.20$ points with task-clustered
bootstrap CI $[-0.80,+1.21]$; pooled cell-level McNemar values summarize the recurring task
IDs across conditions.

\section{Terminal Selection-Rule Ablations}
\label{app:terminal-ablations}

Table~\ref{tab:baselines} tests whether the terminal result depends on the particular
changed-line score used for Text, and separates cross-attempt routing agreement from a
single-candidate routing-concentration score. The surface variants add sensitivity to order,
repetition, token identity, file structure, or patch size. Every rule makes a realized
$1$-of-$K$ choice on the same pools under the main-paper eligibility and grading convention.

\begin{table*}[t]
\centering
\small
\setlength{\tabcolsep}{6pt}
\begin{tabular}{lcc}
\toprule
\textbf{Terminal selector} & \textbf{Macro (\%)} & \textbf{vs.\ \textsc{Risa}} \\
\midrule
Uniform                                  & 44.9 & --- \\
patch-length prior                       & 46.7 & $-1.47$ $[-2.54,-0.37]$ \\
file-set Jaccard                         & 46.9 & $-1.27$ $[-2.38,-0.13]$ \\
changed-line multiset                    & 46.9 & $-1.34$ $[-2.48,-0.17]$ \\
TF--IDF cosine centrality                & 47.5 & $-0.74$ $[-1.71,+0.23]$ \\
character $3$-gram Jaccard               & 47.5 & $-0.70$ $[-1.71,+0.33]$ \\
changed-line Jaccard (Text)              & 48.0 & $-0.20$ $[-1.17,+0.84]$ \\
routing concentration (single candidate) & 45.6 & $-2.64$ $[-3.92,-1.31]$ \\
\midrule
\textbf{\textsc{Risa} routing agreement} & \textbf{48.2} & --- \\
Oracle                                   & 60.9 & --- \\
\bottomrule
\end{tabular}
\caption{Terminal selection-rule ablations on the six \texttt{gpt-oss} conditions
($496$--$498$ eligible tasks each; macro-average of the six). The right column is the paired
difference against \textsc{Risa}, with bootstrap resampling clustered on SWE-bench
task. Surface variants change how patch centrality is measured; routing concentration uses
the identical fingerprint but no cross-attempt comparison. The $2.64$-point separation
identifies cross-attempt agreement, rather than intrinsic concentration, as the useful routing
signal.}
\label{tab:baselines}
\end{table*}

\section{Role Gate: Routing versus Parsed Tool Calls}
\label{app:role-gate}

A natural question is why the write gate reads routing at all, when a candidate ends in a
tool invocation that can be parsed. We replay both gates offline on the recorded candidate
pools, holding everything downstream identical.

\subsection{Agreement}
\label{app:role-agreement}
Over $130{,}464$ candidate actions, the deployed routing gate and an
exact parsed-role gate assign the same write/non-write label $96.8\%$ of the time, and
$99.8\%$ of candidates contain a parseable invocation. The two gates are therefore
near-interchangeable in labeling accuracy.

\subsection{What the Parse Actually Requires}
\label{app:parser-taxonomy}
Write actions span multiple invocation forms: of $54{,}855$ write actions, $97.3\%$ are
shell commands issued through the generic
execution tool (\texttt{sed -i}, a heredoc, \texttt{git apply}, an \texttt{apply\_patch}
invocation), and $2.7\%$ arrive as a dedicated patch tool. The routing gate provides a
portable alternative to a scaffold-specific command taxonomy: it is fit once per model
configuration from the same labels and reads telemetry already produced during inference.

\subsection{Substitution Test}
\label{app:role-substitution}
On the $423$ mixed-outcome write steps of the strong-label
bank, we replace the gate and keep the write score, the cohort statistics, and the
evaluation identical. Selecting within the routing cohort resolves $50.5\%$ of the
step-level branches against a same-step uniform baseline of $49.2\%$ ($+1.3$ points,
step-bootstrap CI $[-2.5,+5.5]$); the parsed cohort gives $+0.7$ points
($[-3.6,+5.0]$) and the ungated cohort $+1.2$ points ($[-3.0,+5.4]$). These aligned
outcomes establish stability across gate implementations; the deployed system uses routing to keep both control levels on one
telemetry source without a scaffold-specific parser.

\section{Write-Score Components}
\label{app:write-score}

The write score $S_{\mathrm{write}}=z(H)+1.5\,z(M)+z(\bar S)-z(P)$ has fixed coefficients,
so its terms can be audited directly. Table~\ref{tab:writecomp} reports step-internal
pairwise AUC on $423$ mixed-outcome steps ($2{,}636$ outcome-labeled candidates), with
step-clustered bootstrap intervals; all statistics are computed exactly as at control time.

\begin{table}[h]
\centering
\small
\begin{tabular}{lc}
\toprule
\textbf{Score} & \textbf{AUC (95\% CI)} \\
\midrule
peer similarity $\bar S$ alone      & \textbf{.558} $[.522,.592]$ \\
mode fraction $M$ alone             & .523 $[.506,.541]$ \\
entropy $H$ alone                   & .512 $[.478,.547]$ \\
concentration $-P$ alone            & .486 $[.450,.521]$ \\
\midrule
deployed $z(H){+}1.5z(M){+}z(\bar S){-}z(P)$ & .520 $[.486,.555]$ \\
equal weights                        & .522 $[.486,.556]$ \\
without $H$                          & .513 $[.477,.547]$ \\
without $M$                          & .518 $[.483,.552]$ \\
without $\bar S$                     & .493 $[.462,.528]$ \\
without $P$                          & .541 $[.507,.574]$ \\
\bottomrule
\end{tabular}
\caption{Components of the write score on the strong-label bank. Mean peer similarity
$\bar S$ provides the clearest standalone signal, and removing it produces the largest degradation among the combination ablations. The $H$ and $P$ terms act
as stability guards; all coefficients were frozen before the campaign and remain fixed across
tasks, models, and effort settings.}
\label{tab:writecomp}
\end{table}

\subsection{Cold-Start Fallback}
\label{app:cold-start}
When a trajectory has no executed history, the exploration rule
falls back to the sibling routing medoid. This deterministic rule fires once per attempt, at
its first step, and only when that step is not already a write step: it never ranks a write
cohort or a final patch. If no fingerprint is available the rule returns the first sampled
candidate, which is exchangeable with a uniform draw because the siblings are sampled
independently from the same prefix.

\section{Pipeline Composition and Sampling-Budget Transfer}
\label{app:pipeline}

Table~\ref{tab:costmatched} evaluates the deployed system end to end and separately tests
whether terminal selection transfers to a much smaller candidate-generation budget. Both
pools use $K{=}4$ separately sampled attempts on the same fixed $200$-task subset
and the same grading convention; they differ only in how each step is generated. The
\emph{steered} pool samples $n{=}16$ candidate actions per step and runs the two-level
controller; the \emph{uncontrolled} pool samples one action per step and executes it, so its
candidate-token cost is roughly $1/16$ of the steered pool at equal $K$ (prefix caching makes
the wall-clock ratio smaller). ``Uncontrolled + Uniform'' is therefore the no-controller
pipeline; steered $+$ Routing is \textsc{Risa}, while steered $+$ Hybrid is \textsc{Risa-H}.

\begin{table}[h]
\centering
\small
\setlength{\tabcolsep}{3.6pt}
\begin{tabular}{llccccc}
\toprule
\textbf{Model} & \textbf{Pool} & \textbf{Unif.} & \textbf{Text} & \textbf{Rout.}
& \textbf{Hyb.} & \textbf{Oracle} \\
\midrule
\texttt{20b} med  & steered      & 45.5 & 49.5 & \textbf{50.5} & 50.0 & 63.5 \\
                  & uncontrolled & 45.4 & 50.0 & 48.0 & \textbf{51.0} & 63.0 \\
\midrule
\texttt{120b} med & steered      & 49.9 & 54.3 & 53.3 & \textbf{55.3} & 65.3 \\
                  & uncontrolled & 53.5 & 56.3 & 54.8 & \textbf{55.3} & 71.9 \\
\bottomrule
\end{tabular}
\caption{Pipeline composition and sampling-budget comparison on the fixed
$200$-task subset (\% resolved,
paper grading convention; $200$ tasks for \texttt{20b} and $199$ with four completed
attempts for \texttt{120b}). End to end, the
\textsc{Risa} (steered $+$ Routing) reaches $50.5\%$ against $45.4\%$ for the
no-controller pipeline (uncontrolled $+$ Uniform) at \texttt{20b}, a task-paired bootstrap
difference of $+5.1$ points, CI $[-0.1,+10.2]$; at \texttt{120b}, \textsc{Risa-H}
(steered $+$ Hybrid)
reaches $55.3\%$ against $53.5\%$ for uncontrolled $+$ Uniform
($+1.8$ points, CI $[-2.6,+6.2]$).
Within the uncontrolled pools, which cost about $1/16$ of the candidate tokens per step,
routing arbitration still adds $2.6$ and $1.3$ points over Uniform and hybrid arbitration
$5.6$ and $1.8$ points, so the terminal level is effective at a small fraction of the
generation budget.}
\label{tab:costmatched}
\end{table}

\section{Cross-Family Reproduction (\texttt{Qwen3.6-35B-A3B})}
\label{app:qwen}

The main grid uses \texttt{gpt-oss-20b} ($24$ layers, $32$ experts, $4$ active) and
\texttt{gpt-oss-120b} ($36$ layers, $128$ experts, $4$ active). To test portability to a
different routing shape, we rerun the pipeline on \texttt{Qwen3.6-35B-A3B} ($40$
layers, $256$ experts, $8$ active), which also has a hybrid attention stack. The
role-gating logic, fixed write coefficients, decision-token fraction, and $K{=}4$ selector
remain
unchanged. Because the raw fingerprint dimension changes, role centroids are re-estimated
from task-disjoint Qwen action labels under the same role definition; the refit uses role
labels exclusively. Sampling follows the vendor's recommended setting
(temperature $0.6$, top-$p$ $0.95$, top-$k$ $20$) rather than the \texttt{gpt-oss}
setting; the step budget, candidate count, and attempt count are unchanged.

\subsection{Full-Benchmark Evaluation}
\label{app:qwen-coverage}
The cross-family condition covers all $500$ SWE-bench Verified tasks, with $K{=}4$
attempts and $n{=}16$ candidates per step. We report it separately from the six
\texttt{gpt-oss} conditions because it uses a different model family and sampling temperature.
Under the main-table requirement of at least two graded patches, $498$ tasks are eligible.
Across the $2{,}000$ scheduled (task, attempt) cells, $28$ produce empty patches, while the
eight cells for \texttt{astropy-8707} and \texttt{astropy-8872} share the reproducible image-build
incompatibility described in Appendix~\ref{app:grading}. Every selector is
computed on the same available patch pool for each eligible task.

\begin{table}[h]
\centering
\small
\setlength{\tabcolsep}{4.5pt}
\begin{tabular}{lcc}
\toprule
\textbf{Selector} & \textbf{Resolved (\%)} & \textbf{Gain vs.\ Uniform} \\
\midrule
Uniform          & 41.7 & --- \\
Text             & 45.0 & $+3.3$ $[+1.7,+4.8]$ \\
\textsc{Risa}   & \underline{45.2} & $+3.5$ $[+1.9,+5.0]$ \\
\textsc{Risa-H} & \textbf{45.6} & $\mathbf{+3.9}$ \\
\midrule
Oracle           & 50.6 & $+8.9$ \\
\bottomrule
\end{tabular}
\caption{Cross-family reproduction: final-patch selection (\% resolved) over $K{=}4$
separately sampled \texttt{Qwen3.6-35B-A3B} attempts on the full $500$-task benchmark
($498$ eligible tasks). Eligibility requires at least two graded patches. Brackets are
task-bootstrap $95\%$ confidence
intervals for paired gains over Uniform. Bold and underlining mark the best and second-best
deployable estimates.}
\label{tab:qwen}
\end{table}

On the full benchmark, \textsc{Risa} improves over Uniform by $3.5$ points
($95\%$ CI $[+1.9,+5.0]$, $p<0.001$), while Text improves by $3.3$ points
($[+1.7,+4.8]$, $p<0.001$). \textsc{Risa} matches Text, with $10$ routing-only wins and
$9$ text-only wins (exact McNemar $p{=}1.000$). \textsc{Risa-H} attains the highest
deployable resolved rate at $45.6\%$. The Uniform-to-Oracle headroom is $8.9$ points, of which \textsc{Risa} recovers
approximately $39\%$.

Among the $476$ tasks with all four attempts graded, resolved rates are $43.0\%$ for
Uniform, $46.4\%$ for Text, $46.6\%$ for \textsc{Risa}, $47.1\%$ for \textsc{Risa-H},
and $52.1\%$ for Oracle. \textsc{Risa} exceeds Uniform by $3.6$ points
($95\%$ CI $[+2.0,+5.3]$), and the \textsc{Risa}--Text McNemar count remains $10{:}9$
($p{=}1.000$).

\subsection{Generation Outcome}
\label{app:qwen-outcome}
The condition records $1{,}964$ completed (task, attempt) cells out of $2{,}000$ scheduled,
with a $73.5\%$ patch yield. Stored diffs preserve the
accumulated patch at budget termination. Grading these final patches yields a $42.1\%$
attempt-level resolved rate, alongside $0.1\%$ under stock exit-triggered scoring.
Table~\ref{tab:qwen} reports task-level selector outcomes from the same re-graded patch pool.

\section{Hyperparameters}
\label{app:hyperparameters}

\begin{table}[h]
\centering
\small
\begin{tabular}{p{0.32\columnwidth}p{0.55\columnwidth}}
\toprule
parameter & value \\
\midrule
attempts per task $K$ & 4 \\
candidates per step $n$ & 16 \\
sampling temperature & 1.0, top-$p$ 1.0 (\texttt{gpt-oss}); 0.6, top-$p$ 0.95, top-$k$ 20 (\texttt{Qwen3.6}) \\
decision-token fraction & lowest 25\% of patch tokens by teacher-forced probability \\
step fingerprint span & tool-call segment of the candidate (fallback: trailing 64 generated tokens) \\
\texttt{gpt-oss} reasoning efforts & native low / medium / high setting \\
max new tokens per query & 4096 (low/medium), 65536 (high) \\
step budget per attempt & 50 agent steps \\
history pool (exploration rule) & last 64 executed actions; the $\min(3,|\mathcal H_t|)$ most similar are scored \\
write-rule cluster threshold & $\tau = 0.65$ (pairwise WJ) \\
write-rule score & $z(H)+1.5\,z(M)+z(\bar S)-z(P)$; coefficients fixed at $(1,1.5,1,-1)$ \\
\texttt{Qwen3.6} generation budget & 8192 new tokens per query, same 50-step budget \\
top-$k$ logprobs requested & 20 per generated token \\
role-centroid fit & task-disjoint 70/30 split, nearest-centroid (cosine) \\
terminal fingerprint prompt & problem statement $+$ accumulated diff, \texttt{max\_tokens}${=}1$, temperature $0$ \\
bootstrap resamples & $10^4$ (task-clustered for pooled cells) \\
agent container limits & 4 CPUs, 2\,GB RAM \\
\bottomrule
\end{tabular}
\caption{Final configuration. All arms of an ablation share every value except the
selection rule under study.}
\label{tab:hyperparameters}
\end{table}

\subsection{Selection of Final Settings}
\label{app:setting-selection}
Three groups of knobs
exist, and each was set by a stated criterion rather than by tuning on evaluation outcomes.
(i)~\emph{Budget parameters} ($K{=}4$ attempts, $n{=}16$ candidates, $50$ steps) were fixed
in advance according to the compute available for a full $500$-task grid at three effort
settings. (ii)~\emph{Routing-readout parameters} were
selected on diagnostic data before the campaign, then swept afterwards as robustness checks:
decision-token fraction over $\{5,10,25,50,100\}\%$, write-rule threshold
$\tau\in\{0.55,0.65,0.75,0.85\}$, exploration neighborhood $m\in\{1,3,5,10\}$ crossed with
history pool $\in\{8,64\}$, and fingerprint window over
$\{$whole span, opening $64$, trailing $64$, decision tokens$\}$. The criterion was the
diagnostic pairwise AUC on outcome-labeled steps and attempt pairs, which is disjoint from
the evaluation grid; Table~\ref{tab:sens} shows every swept value.
(iii)~\emph{Write-score coefficients} $(1,1.5,1,-1)$ were fixed once on the separate
$98$-step mixed-outcome pilot (Section~\ref{app:write-rule}) and remain fixed throughout
evaluation. All final parameters follow these diagnostic and pilot criteria, independently
of the resolved rates reported in the main table.

\section{Compute Infrastructure and Cost}
\label{app:compute}

All experiments ran on a single server, shared with other users, with
8$\times$ NVIDIA H20 (95.6\,GiB each, compute capability 9.0), two Intel Xeon Platinum
8480+ CPUs ($56$ cores / $112$ threads each, $224$ logical cores) and $2.0$\,TB of system
RAM, under Ubuntu 22.04.5 LTS (Linux 5.15.0) with NVIDIA driver 560.35.05 (CUDA 12.6) and
Docker 27.5.1. The serving stack is vLLM 0.15.1 with PyTorch 2.9.1+cu128, Transformers
4.57.3 and Python 3.12.9; the agent and all analysis scripts run on the same Python
version with NumPy, pandas and the official \texttt{swebench} harness. The 20B model
is served with tensor parallelism 1 (one replica per GPU, data-parallel round-robin);
the 120B model with tensor parallelism 2 (three GPU pairs); the cross-family
\texttt{Qwen3.6-35B-A3B} condition uses eight tensor-parallel-1 replicas. SWE-bench
instance images are built and graded locally in Docker, with each evaluation container
limited to 8 CPUs and 16\,GB RAM.

\subsection{Cost of the Method}
\label{app:method-cost}
The $n{=}16$ step candidates share a prefix and are
issued as a batch, so prefix caching reduces repeated prefill work. On our serving stack the
measured prefix-cache hit rate is $98.9\%$, and the realized wall-clock latency on the
same hardware for an $n{=}16$ step is $\approx 2\times$ that of a single-sample step. A
medium-effort steered rollout generates $\approx 90$K tokens including all candidate
continuations ($\approx 360$K generated
tokens per task at $K{=}4$). Per-step fingerprint scoring is arithmetic over returned tensors; terminal selection
adds one short prefill and one discarded token per available patch, with no additional
patch generation. Representative measured
generation costs (wall-clock $\times$ devices) are 22~GPU-h (\texttt{20b} low),
222~GPU-h (\texttt{20b} high), 25~GPU-h (\texttt{120b} low), and $\approx 208$~GPU-h
for a 200-task \texttt{120b}-high subset served as three 2-GPU replicas. These values provide
representative reference points; high-effort cost is dominated by
long-context re-prefill rather than decode.
\subsection{Randomness and Seeds}
\label{app:randomness}
Two sources of randomness exist. Trajectory sampling uses the inference server's stochastic
sampler: each of the $K$ attempts is an
independent draw at the configured temperature, which is the intended behavior, since the
method operates on independently sampled attempts. Consequently single-attempt numbers vary
between runs, and every reported comparison is either paired on the same stored pool (all
selector rows) or aggregated over $K$ attempts and hundreds of tasks. All post-hoc
randomness is seeded and reproducible from the released code: bootstrap resampling uses
\texttt{numpy.random.default\_rng(0)}, trajectory subsampling for step statistics uses
\texttt{random.Random(0)}, the diagnostic-pool construction and the $70/30$ centroid split
use \texttt{seed}${=}42$; the corresponding diagnostic-pool and split id lists are released
with the code, so the exact task sets can be reproduced without rerunning the sampler.

\subsection{Evaluation Metrics}
\label{app:evaluation-metrics}
The primary metric is the official SWE-bench Verified
\emph{resolved rate}: a patch counts as resolved when the instance's fail-to-pass and
pass-to-pass test sets both pass inside the instance container, as decided by the benchmark
harness. This community-standard criterion grounds evaluation in execution rather than
surface similarity to a reference patch. A selector's score is the resolved rate of the single patch it submits
per task, so all selectors are compared as realized $1$-of-$K$ decisions rather than as
rankings. Three reference quantities frame those numbers: \emph{Uniform}, the exact
expectation of choosing uniformly among a task's available patches; \emph{Oracle}, the
union of successes in the pool, an upper bound on what any selector could reach; and
\emph{patch yield}, the fraction of attempts that end with a nonempty diff, which separates
candidate availability from candidate choice. For diagnostic analyses on labeled steps and
attempt pairs we use pairwise AUC --- the probability that a resolving branch is ranked
above a non-resolving one --- because those comparisons are about ordering candidates, not
about absolute calibration. Macro-averages weight each model--effort condition equally so
that the six conditions, which differ in difficulty, contribute equally.

\section{Additional Pilot Measurements}
\label{app:pilot}

\begin{table*}[t]
\centering
\small
\setlength{\tabcolsep}{6pt}
\begin{tabular}{lccccc}
\toprule
decision-token fraction & 5\% & 10\% & 25\% & 50\% & 100\% \\
\midrule
offline AUC (cross-attempt) & .644 & .648 & \textbf{.657} & .657 & .639 \\
realized macro-avg (\%) & --- & 48.0 & \textbf{48.2} & 48.2 & --- \\
\midrule
\multicolumn{6}{l}{\emph{per-condition realized spread across fractions: at most
$2.4$ points; the selected 25\% setting attains the top macro-average.}} \\
\addlinespace
\multicolumn{6}{l}{write-rule threshold $\tau \in \{.55,.65,.75,.85\}$:
pairwise AUC .548/.572/.536/.580} \\
\multicolumn{6}{l}{step-level fraction $\{10,25,50\}\%$: peer-similarity AUC
.554/.545/.544} \\
\multicolumn{6}{l}{exploration $m\in\{1,3,5,10\}$, pool $8$: .546/.545/.548/.551;
pool $64$: .533/.537/.547/.551} \\
\bottomrule
\end{tabular}
\caption{Sensitivity of the selected routing-window and neighborhood hyperparameters.
Candidate count, attempt count, step budget, and the four write-rule coefficients are held
fixed throughout.}
\label{tab:sens}
\end{table*}

\subsection{Window-Control Battery}
\label{app:window-controls}
Using each patch's mean WJ agreement with the
other available attempts as its score, AUC for ranking a resolved patch above an unresolved
one in mixed-outcome groups is, by readout window: decision tokens (lowest-probability
25\%) $.657$; whole span $.639$; random 25\% (3 seeds) $.634$; uniformly spaced 25\%
$.620$; router-weight entropy in place of token probability $.620$; highest-probability
25\% (complement) $.616$. Decision tokens lead the whole-span readout by $.018$ AUC and
the probability-complement control by $.041$.

\subsection{Routing Saturation}
\label{app:saturation}
The fraction of layer--expert cells receiving nonzero routing
mass rises from $0.75$ (spans under 64 tokens) to $0.96$ (beyond a thousand). Identity AUC, which uses WJ to distinguish pairs from the same task-and-prefix generation
context from pairs drawn across contexts, falls to $.58$ for the trailing-64 window on long
spans,
while whole-span, decision-token, and
opening-64 readouts retain $.88$, $.93$, and $.98$, respectively.

\subsection{Choice Concentration and Cross-Attempt Agreement}
\label{app:decision-concentration}
The lowest-probability quartile
of final-step generations carries a median $72\%$ of total token surprisal
$\sum_t-\log p_t$ ($n{=}658$ generations). Mean cross-attempt routing agreement is
$0.56$ at the remaining higher-probability positions and $0.65$ at decision tokens; the
main-paper decile plot reports the full distribution.

\subsection{Threshold Sensitivity}
\label{app:threshold-sensitivity}
The selected routing-window and neighborhood
hyperparameters occupy a broad stable region (Table~\ref{tab:sens}). Fractions from
10--50\% are stable both offline (cross-attempt ranking AUC) and \emph{realized}
(resolved rate when the terminal selector uses each fraction, macro-averaged over the six
model--effort conditions); the selected $25\%$ fraction attains the top offline AUC and
realized macro-average. Across the threshold sweep, pairwise AUC remains in the compact
$.536$--$.580$ range.

\subsection{Novelty Across Steps and Trajectories}
\label{app:novelty-shape}
We bin candidates by the similarity of their routing fingerprint to the trajectory's own
recent executed actions, using the deployed score (mean of the three most similar entries
in a $64$-action history) and, as a robustness check, the mean over the whole history.

\begin{figure}[h]
\centering
\includegraphics[width=0.48\columnwidth]{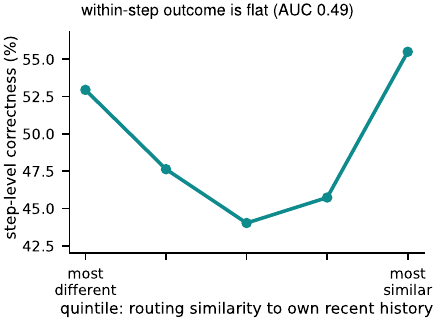}\hfill
\includegraphics[width=0.48\columnwidth]{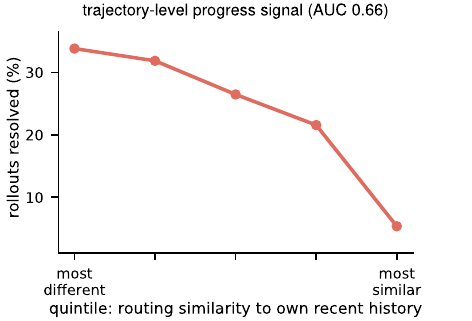}
\caption{Similarity to one's own recent routing history. Left: within-step outcome balance
supports the diversity-prior interpretation ($n{=}2{,}636$). Right: across rollouts, mean history similarity strongly separates
resolved from unresolved attempts ($n{=}1{,}021$), motivating a trajectory-level
diversity prior.}
\label{fig:novelty}
\end{figure}

Within a step, routing-history quintiles yield pairwise AUCs near chance, identifying history
disagreement as a computation-diversity prior.
Over $2{,}636$ branch-labeled write candidates from mixed-outcome steps, the resolving
fraction (most different first) is $52.9$, $47.6$, $44.0$, $45.7$, $55.5$ percent under the deployed
score (pairwise AUC $0.49$) and $51.6$, $48.6$, $50.9$, $48.4$, $46.4$ under the
whole-history mean (AUC $0.52$). Figure~\ref{fig:novelty} (left) visualizes this
within-step outcome balance.

Across trajectories, the relationship is strong and stable under both scores. Binning
$1{,}021$ rollouts by mean history similarity gives resolve rates of $33.8$, $31.9$,
$26.5$, $21.6$, $5.4$ percent (AUC $0.656$; $30.4$ to $11.7$ percent, AUC $0.611$, under
the whole-history mean). Thus routing-history similarity is a strong progress marker
(Figure~\ref{fig:novelty}, right). The controller operationalizes this stable trajectory-level signal as a diversity prior;
targeted intervention studies can further characterize how actively changing routing
history affects downstream outcomes.

\subsection{Steering Coverage Analysis}
\label{app:steering-coverage}
On the $80$-task empirically hard diagnostic
pool, we compare the full controller with the same agent scaffold and controller disabled.
A submittable patch is an attempt ending with a nonempty final diff. Routing-guided steering
raises this attempt-level yield from $79\%$ to $94\%$, reducing patchless attempts from
$21\%$ to $6\%$. This coverage diagnostic measures candidate availability: the controller
turns more otherwise stalled runs into candidates that terminal selection can compare. Task-level
selector accuracy is reported separately on fixed steered pools in the main paper.

\section{Reproducibility Statement}
\label{app:reproducibility}

Code for the agent scaffold, serving-side routing capture, selection rules, grading pipeline,
and analysis scripts will be released under a research-friendly license upon publication.
SWE-bench Verified is public, and the released pipeline regenerates per-candidate routing
traces (expert indices and gate weights). The release will include regeneration scripts,
derived pilot tables, and trace artifacts distributed under their source licenses. Analysis
scripts fix their random seeds; trajectories use independent stochastic draws at the
model-specific temperatures in Table~\ref{tab:hyperparameters}. Reported comparisons
aggregate over $K$ attempts and use paired arms and instances where applicable.

\end{document}